\documentclass[lettersize,journal]{IEEEtran}
\usepackage{amsmath,amsfonts}
\usepackage{algorithmic}
\usepackage{algorithm}
\usepackage{array}
\usepackage[caption=false,font=normalsize,labelfont=rm,textfont=rm]{subfig}
\usepackage{textcomp}
\usepackage{stfloats}
\usepackage{url}
\usepackage{verbatim}
\usepackage{graphicx}
\usepackage{cite}
\usepackage{multirow}

\begin{document}

\title{Kernel Stochastic Configuration Networks for Nonlinear Regression}

\author{Yongxuan Chen,~\IEEEmembership{}Dianhui Wang~\IEEEmembership{}

        % <-this % stops a space
\thanks{\textit{Corresponding author:}\textbf{ dh.wang@deepscn.com}}% <-this % stops a space
%\thanks{Manuscript received April 19, 2021; revised August 16, 2021.}
}

% The paper headers
\markboth{}%
{Shell \MakeLowercase{\textit{et al.}}: A Sample Article Using IEEEtran.cls for IEEE Journals}

%\IEEEpubid{0000--0000/00\$00.00~\copyright~2021 IEEE}
% Remember, if you use this you must call \IEEEpubidadjcol in the second
% column for its text to clear the IEEEpubid mark.

\maketitle

\begin{abstract}
Stochastic configuration networks (SCNs), as a class of randomized learner models, are featured by its way of random parameters assignment in the light of a supervisory mechanism, resulting in the universal approximation property at algorithmic level. This paper presents a kernel version of SCNs, termed KSCNs, aiming to enhance model’s representation learning capability and performance stability. The random bases of a built SCN model can be used to span a reproducing kernel Hilbert space (RKHS), followed by our proposed algorithm for constructing KSCNs. It is shown that the data distribution in the reconstructive space is favorable for regression solving and the proposed KSCN learner models hold the universal approximation property. Three benchmark datasets including two industrial datasets are used in this study for performance evaluation. Experimental results with comparisons against existing solutions clearly demonstrate that the proposed KSCN remarkably outperforms the original SCNs and some typical kernel methods for resolving nonlinear regression problems in terms of the learning performance, the model’s stability and robustness with respect to the kernel parameter settings. 
\end{abstract}


\section{Introduction}
\IEEEPARstart{N}{EURAL} networks have been widely applied for nonlinear regression analysis, aiming at establishing a feasible representation between observed predictors and responses through learning from a collection of samples [1]-[3]. To develop faster learner models, researchers have introduced randomized algorithms for training neural networks [4], [5]. The idea behind these randomized learning techniques lies in the random assignment of input weights and biases followed by solving linear regression problem for evaluating the output weights. However, the existing randomized algorithms lack basic logic in use [6]. In 2017, Wang and Li innovatively proposed an advanced randomized learner model, termed stochastic configuration networks (SCNs) [7], which can be incrementally constructed in light of a supervisory mechanism. A remarkable difference between the SCN framework and other randomized learner models is the guaranteed universal approximation property at algorithmic level, that is, SCNs can ensure zero-free learning performance as the random nodes keep being added. Due to its theoretical contribution and practiced value for data analytics, SCNs has received considerable attention in these years [8]-[16]. Theoretically, an SCN model with enough random nodes can provide arbitrarily good approximations to a prescribed function. In practice, with the number of random nodes continuously increasing, the risk of overfitting situation is accordingly raised. How to enhance the representation learning abilities of hidden nodes is worth investigating. On the other hand, as a randomized learner model, different forms of configured parameters have great impacts on the performance stability of an SCN model. Thus, to develop an improved SCN framework with better representation learning capability and performance stability is of great necessity for data analytics.
\begin{figure}[!t]
\centering
\subfloat[]{\includegraphics[width=2.5in]{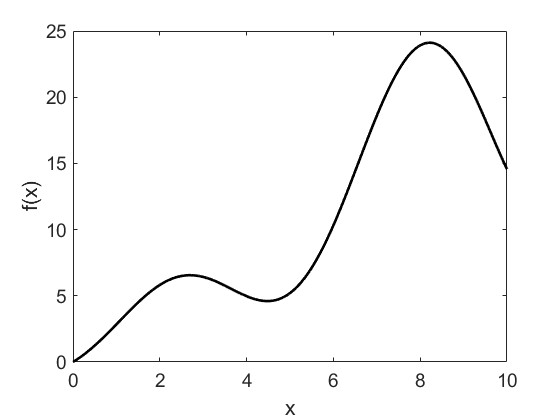}%
\label{fig_first_case}}
\hfil
\subfloat[]{\includegraphics[width=2.5in]{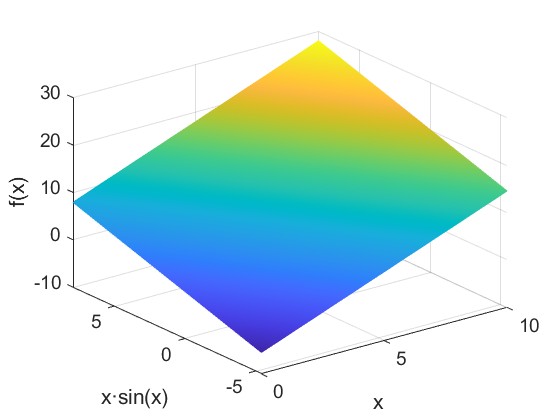}%
\label{fig_second_case}}
\caption{An illustration of deducting nonlinearities by high-dimensional mapping: (a) the function plot of \(f(x)=2x+xsin(x)\) where \( x\in[0,10]\). (b) the function plot of \(f[\phi(x)]=[2\; 1]\cdot\phi(x)\) where \(\phi(x)=[x\; xsin(x)]^T\).}
\label{fig_sim}
\end{figure}

In machine learning theory, kernel techniques have demonstrated their strong power in terms of enhancing the representation learning ability at algorithmic level [17]-[19]. The concept of kernel-based approaches lies in the high-dimensional mapping for tackling nonlinearities among original data space, which can be illustrated in Fig. 1. The most representative kernel-based learner models for nonlinear regression are support vector regression (SVR) [20], [21] and radial basis function networks (RBFN) [22]. One of the most desirable features about these approaches is the powerful nonlinear expression ability in regression tasks. However, the vast majority of the existing kernel-based learner models, including SVR and RBFN, seeks to project the original data onto a feature space for tackling nonlinearities. The effectiveness of this operation greatly depends on the specific form of the kernel function, which defines a unique reproducing kernel Hilbert space (RKHS), due to the unknown distribution of training data. Thus, the performances of these models are sensitive to kernel parameter settings, which brings difficulties to establish a reliable learner model with stable performances. A learner model’s robustness with respect to the kernel parameter settings is helpful for improving its generalization capability and training efficiency.
\begin{figure}[!t]
\centering
\includegraphics[width=2.5in]{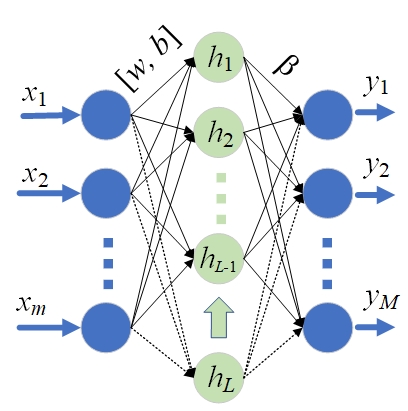}
\caption{The schematic of SCNs.}
\label{fig_1}
\end{figure}

This paper aims to develop an applicable nonlinear learner model by exploring the kernel version of SCNs, termed kernel SCNs (KSCNs), for resolving complex nonlinear regression problems. A remarkable difference between the proposed KSCNs and other traditional kernel-based learner models lies in the way of constructing the high-dimensional feature space. In KSCNs, the high-dimensional mapping is based on the random bases of SCNs which are generated under the supervisory mechanism. On the other hand, by introducing the kernel technique, the representation learning ability of the model is enhanced compared to original SCNs. The experimental results show that the proposed method demonstrates superior performances over SCNs and some representative kernel-based learner models, in terms of the learning performance, the model’s stability and robustness with respect to the kernel parameter settings. The contributions of this article can be summarized as follows.

1) The kernel SCNs (KSCNs) framework is proposed by establishing the connection between a RKHS and a high-dimensional feature space where the random bases are mapped onto.

2) The convergence analysis of KSCNs is provided to demonstrate that the proposed KSCNs hold the universal approximation property.

3) An early stopping strategy is designed at the training stage of KSCNs for preventing the overfitting situation and improving the modeling efficiency.

4) Good illustration of applications, including two actual industrial case studies, of KSCNs is demonstrated with detailed performance evaluations.

The rest of this article is organized as follows: Section II briefly introduces some necessary preliminaries. Section III details our proposed KSCNs. Section IV provides three case studies to verify the effectiveness of the proposed method. Section V concludes this article.

\section{Preliminaries}
This section provides the necessary preliminaries of this paper, which are the brief introductions of SCNs and RKHS.
\subsection{Stochastic Configuration Networks}

Stochastic configuration networks (SCNs) [7] belong to the randomized learner model with universal approximation property. Unlike the conventional randomized learner model such as random vector functional-link (RVFL) [23] networks, SCNs randomly configure their model parameters through a well-designed supervisory mechanism. Furthermore, by avoiding backpropagation mechanism, the training procedures of SCNs are computationally lightweight and timesaving, which is suitable for nonlinear data modeling. The schematic of SCNs can be found in Fig. 2.

Given data matrix \(X=\left[x_1,x_2,\ldots,\ x_n\right]^T\in\mathbb{R}^{n\times m}\) and \(Y=\left[y_1,y_2,\ldots,y_n\right]^T\in\mathbb{R}^{n\times M}\). Assuming an SCN has been constructed with \(L-1\) hidden nodes, which is \(f_{L-1}\left(X\right)=H_{L-1}\beta=\sum_{j=1}^{L-1}{\beta_jh_j(X,\omega_j,b_j)}\) where \(h_j\left(X,\omega_j,b_j\right)=\left[g\left(\omega_j^Tx_1+b_j\right),\ldots,g\left(\omega_j^Tx_n+b_j\right)\right]^T\) (\(\omega_j\) and \(b_j\) are the input weights and biases of the \(j\)th hidden node and \(g\) is the activation function). Denoting the residual error for the current SCN is \(e_{L-1}=f-f_{L-1}=\left[e_{L-1,1},\ldots,e_{L-1,M}\right]\), where \(f\) is the objective function. If \(\|e_{L-1}\|\) does not meet the required tolerance level, then a new node \(h_L\left(X,\omega_L,b_L\right)\) is expected to be generated to the current model.

The \(h_L\) is selected to satisfy the following inequalities:
\begin{equation}
\label{deqn_ex1a}
〈e_{L-1,q},h_L 〉^2\ge b_g^2 \delta _{L,q},q=1,2,…,M
\end{equation}
where \(0<\|h_L \|<b_g\) for some \(b_g\in\mathbb{R}^{+}\) and \(\delta_{L,q}=(1-r-\mu_L)\|e_{L-1,q} \|^2\) for \(0<r<1\) and { \(\{\mu_{L}\}\) } is a non-negative real number sequence with \(\underset{L\rightarrow+\infty}\lim{\mu_L}=0\). The above inequalities guarantee that SCNs are with the universal approximation property.

After selecting enough nodes of SCNs incrementally, the output weights of SCNs can be obtained by solving the global least squares problem as:
\begin{equation}
\label{deqn_ex1a}
[\beta_1,\beta_2,…,\beta_L ]=argmin_\beta \|f-\sum_{j=1}^L \beta_j h_j \|.
\end{equation}

\subsection{Reproducing Kernel Hilbert Space}
A positive definite kernel \(K(x,y)\), where \(x\) and \(y\) are the subsets of a compact set \(X\subset R^N\), is the symmetric function of two variables satisfying the Mercer theorem [17], which can define a Hilbert space \(\mathcal{H}\).

If a kernel \(K(x,y)\) has the following property:
\begin{equation}
\label{deqn_ex1a}
f\left(y\right)=\left\langle f\left(x\right),K\left(x,y\right)\right\rangle_\mathcal{H},\ \forall f\in\mathcal{H}
\end{equation}
where \(\left\langle.,.\right\rangle_\mathcal{H}\) denotes the dot product in \(\mathcal{H}\), then \(K(x,y)\) is defined as a reproducing kernel for \(\mathcal{H}\) and \(\mathcal{H}\) is accordingly called a reproducing kernel Hilbert space.

According to Mercer’s theorem, the reproducing kernel \(K(x,y)\) can be decomposed in the form of
\begin{equation}
\label{deqn_ex1a}
K\left(x,y\right)=\sum_{i=1}^{m}{\lambda_i\phi_i\left(x\right)\phi_i\left(y\right),\ m\le\infty}
\end{equation}
where \(\phi_i(.)\) is the eigenfunction and \(\lambda_i\) is the corresponding eigenvalue.

Assuming a nonlinear mapping \(\Phi\) maps \(x\) onto a high-dimensional feature space \(\mathcal{F}\) as 
\begin{equation}
\begin{split}
\label{deqn_ex1a}
&\Phi\left(x\right)=\left(\sqrt{\lambda_1}\phi_1\left(x\right),\sqrt{\lambda_2}\phi_2\left(x\right),\ldots,\sqrt{\lambda_m}\phi_m\left(x\right)\right)\\&x\in\mathcal{X}\rightarrow\Phi\left(x\right)\in\mathcal{F}.
\end{split}
\end{equation}

Then we can rewrite (4) as
\begin{equation}
\label{deqn_ex1a}
K\left(x,y\right)=\sum_{i=1}^{m}{\sqrt{\lambda_i}\phi_i\left(x\right){\sqrt{\lambda_i}\phi}_i\left(y\right)}={\Phi\left(x\right)}^T\Phi\left(y\right).
\end{equation}

The formulation above builds a straightforward connection between a RKHS and the feature space \(\mathcal{F}\). That indicates that the dot product in feature space \(\mathcal{F}\) can be calculated by kernel \(K\left(x,y\right)\) without knowing the specific form of \(\phi_i(.)\), which is also called the kernel trick [17].

\begin{figure}[!t]
\centering
\includegraphics[width=3in]{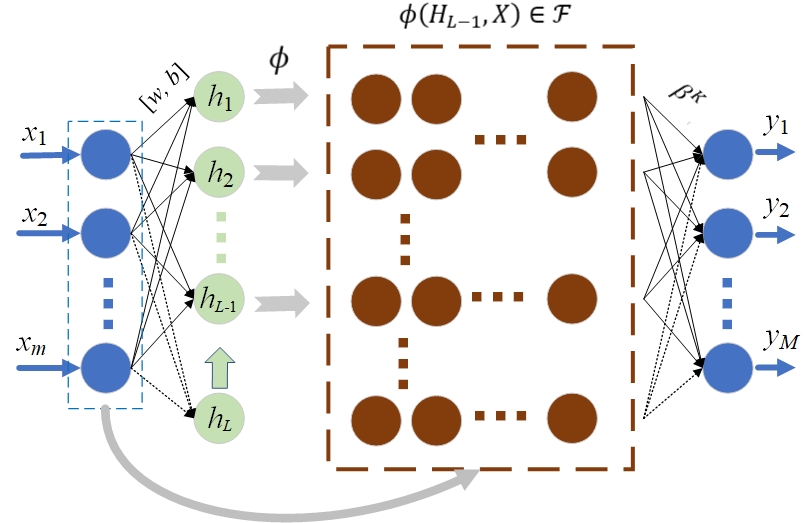}
\caption{The schematic of KSCNs.}
\label{fig_1}
\end{figure}

\section{Kernel Stochastic Configuration Networks}
The random bases of SCNs are generated based on the supervisory mechanism, which are data-dependent and quality-oriented. This paper utilizes these random bases to span an RKHS for deducting data nonlinearities. The universal approximation property is guaranteed based on the SCNs framework and the superiority of the proposed method is explained through investigating the eigenvalue decomposition of kernel Gram matrix. This section details the proposed KSCNs framework and its algorithm descriptions.
\subsection{Construction of KSCNs}
Assuming an SCN with \(L-1\) (\(L=1,2,...\)) hidden nodes has been built, which is \(f_{L-1}\left(X\right)=H_{L-1}\beta=\sum_{j=1}^{L-1}{\beta_jh_j\left(X,\omega_j,b_j\right)}\) where \(h_j\left(X,\omega_j,b_j\right)=\left[g\left(\omega_j^Tx_1+b_j\right),\ldots,g\left(\omega_j^Tx_n+b_j\right)\right]^T\) and \(f_0=0\). To deduct the strong nonlinearities among hidden layer outputs and target variables, a high-dimensional nonlinear mapping \(\phi\) is introduced to project the hidden layer outputs \(H_{L-1}\) and the original input data \(X\) onto a feature space \(\mathcal{F}\) as \(\phi\left(H_{L-1},X\right)\in\mathcal{F}\) where \(H_{L-1}=\left[h_1,\ldots{,h}_{L-1}\right]\). Therefore, the formulation of the current model is \(f_{L-1}^K\left(X\right)=\phi (H_{L-1},X)\beta^K\) where \(\beta^K\) denotes the output weights.

The output weights \(\beta^K\) of KSCNs can be obtained by solving the least squares problem between hidden layer outputs \({\phi(H}_{L-1}^K,X)\) and target variables \(Y\) as 
\begin{equation}
\begin{split}
\label{deqn_ex1a}
\beta^K{=argmin}_{\beta^K}\ \frac{1}{2}\|Y-\sum_{j=1}^{L-1}{\beta_j^K\phi}\|^2+\frac{1}{2}\tau\|\beta^K\|^2
\end{split}
\end{equation}
where \(\tau\) is the regularization factor to prevent the model from falling into the overfitting situation. A closed-form solution to (7) is
\begin{equation}
\begin{split}
\label{deqn_ex1a}
\beta^K={\phi^T\left(\phi\phi^T+\tau I\right)}^{-1}Y
\end{split}
\end{equation}
where \(I\) is the identify matrix. Therefore, the output values of the current model are \(f_{L-1}^K={\phi\beta}^K=\phi{\phi^T\left(\phi\phi^T+\tau I\right)}^{-1}Y\). 

Here, the specific form of high-dimensional mapping \(\phi\) is unknown, which makes (8) cannot be calculated directly. According to (6), it is indicated that the dot product in feature space \(\mathcal{F}\) can be expressed by the kernel function as \(K\) whose calculation can be completed by the commonly used Gaussian kernel function as
\begin{equation}
\begin{split}
\label{deqn_ex1a}
K_{i,j}&=\phi\left(H_{L-1}\left(i\right),x_i^T\right){\phi\left(H_{L-1}\left(j\right),x_j^T\right)}^T\\&=\exp(-\frac{\|{[H}_{L-1}(i),x_i^T]-[H_{L-1}(j),x_j^T]\|^2}{c})
\end{split}
\end{equation}
where \(c\) is the predefined kernel parameter and \(K_{i,j}\) represents the element in \(i\)th row and \(j\)th column of kernel Gram matrix \(K\) and \(H_{L-1}\left(i\right)=[h_1\left(x_i\right),\ldots,h_{L-1}\left(x_i\right)]\) represents the \(i\)th row of \(H_{L-1}\). Therefore, the outputs of the current model can be rewritten as
\begin{equation}
\begin{split}
\label{deqn_ex1a}
f_{L-1}^K=\phi{\phi^T\left(\phi\phi^T+\tau I\right)}^{-1}Y=K\left(K+\tau I\right)^{-1}Y.
\end{split}
\end{equation}

Denoted the residual error produced by the current model as \(e_{L-1}^K=f-f_{L-1}^K=\left[e_{L-1,1}^K,e_{L-1,2}^K,\ldots,e_{L-1,M}^K\right]\). If \(\|e_{L-1}^K\|\) does not achieve the tolerance level or \(L\le L_{max}\) where \(L_{max}\) is the predefined maximum number of hidden nodes, then a new node \(h_L\) is expected to be generated to the model. The generation of \(h_L\) must satisfy some conditions to guarantee the universal approximation property, which is detailed in the Theorem 1. The schematic of KSCNs is demonstrated in Fig. 3.

For unknown testing data matrix \(X_t\), it is also needed to execute the nonlinear mapping process and the testing kernel matrix \(K_t\) is calculated as
\begin{equation}
\begin{split}
\label{deqn_ex1a}
K_{t\left(i,j\right)}&=\phi\left(H_{L-1}\left(i\right),x_i^T\right){\phi_t\left(H_{t,L-1}\left(j\right),x_{t,j}^T\right)}^T\\&=\exp(-\frac{\|{[H}_{L-1}(i),x_i^T]-[H_{t,L-1}(j),x_{t,j}^T]\|^2}{c})
\end{split}
\end{equation}
where \(K_{t(i,j)}\) represents the element in \(i\)th row and \(j\)th column of testing dataset kernel Gram matrix \(K_t\), and \(H_{t,L-1}(j)=[h_{t,1}\left(x_{t,j}\right),\ldots,h_{t,L-1}\left(x_{t,j}\right)]\) represents the \(j\)th row of \(H_{t,L-1}\).

Therefore, the prediction values produced by the current model is given by
\begin{equation}
\begin{split}
\label{deqn_ex1a}
f_{t,L-1}^K=\phi_t{\phi^T\left(\phi\phi^T+\tau I\right)}^{-1}Y=K_t\left(K+\tau I\right)^{-1}Y.
\end{split}
\end{equation}

\subsection{Convergence Analysis}
One of the most desirable properties of SCNs is the universal approximation property. Generally, one cannot expect a learner model to be with good generalization but poor learning performance [7]. Hence, this section produces the universal approximation property of KCSNs.

\textbf{\textit{Theorem 1:}} Suppose that span(\(\Gamma\)) is dense in \(L_2\) space and \(\forall h\in\Gamma\), \(0<\|h\|<b_h\) for some \({b_h\in\mathbb{R}}^+\). Given \(0<r<1\) and a non-negative sequence \(\{\mu_L\}\), \(\mu_L\le\left(1-r\right)\) and \(\underset{L\rightarrow+\infty}\lim{\mu_L=0}\). For \(L=1,2,...,\) denoted by
\begin{equation}
\begin{split}
\label{deqn_ex1a}
\delta_L^K=\sum_{q=1}^{m}\delta_{L,q}^K, \delta_{L,q}^K=\left(1-r-\mu_L\right)\|e_{L-1,m}^K\|^2
\end{split}
\end{equation}
where \(m=1,2,...M\). If \(h_L\) satisfied the following inequality
\begin{equation}
\begin{split}
\label{deqn_ex1a}
{\langle e_{L-1,m}^K,h_L \rangle }^2 \geq b_h^2\delta_L^K, m=1,2,...M
\end{split}
\end{equation}
and the output weights \(\beta^K\) are calculated by (7). Then, we can obtain  \(\underset{L\rightarrow+\infty}\lim{\|f-f_L^K\|=0}\), where \(f_L^K=\phi\left(H_L,X\right)\beta^K=\sum_{j=1}^{L}{\beta_j^lh_j}+\sum_{j=1}^{D-L}{\phi_j\left(H_L,X\right)\beta_j^{nl}}\) where \(D\) is the implicit data dimension in high-dimensional space.

\textit{Proof: } Define intermediate values \(\beta_{L,q}^\prime=\langle e_{L-1,q}^K,h_L \rangle /\|h_L\|^2\) (\(q=1,2,...,M\)) and \(\beta_L^\prime=\left[\beta_{L,1}^\prime,\beta_{L,2}^\prime,\ldots,\beta_{L,M}^\prime\right]^T\), \(e_L^\ast=f-\sum_{j=1}^{L}{\beta_j^\ast h_j=\left[e_{L,1,}^\ast e_{L,2}^\ast,\ldots,e_{L,M}^\ast\right]}\) and \(\left[\beta_1^\ast,\beta_2^\ast,\ldots,\beta_L^\ast\right]={argmin}_\beta\|f-\sum_{j=1}^{L}{\beta_jh_j}\|\). Since the identity-mapping is the specific form of nonlinear mapping \(\phi\), it is obvious to obtain \(\|e_L^K\|^2\le\|e_L^\ast\|^2\) due to the reason that 
\begin{equation}
\begin{split}
\label{deqn_ex1a}
\|e_L^K\|^2=&\|f-\sum_{j=1}^{L}{\beta_j^lh_j}-\sum_{j=1}^{D-L}{\phi_j\left(H_L,X\right)\beta_j^{nl}}\|^2\\\le &\|f-\sum_{j=1}^{L}{\beta_j^\ast h_j}\|^2=\|e_L^\ast\|^2.
\end{split}
\end{equation}

With the conclusions of [7] and above, it can be derived that \(\|e_L^K\|^2=\|e_{L-1}^K-\phi_L\left(H_L,X\right)\beta_L^K\|^2\le\|e_{L-1}^K-\beta_L^\ast h_L\|^2\le \|e_{L-1}^K-\beta_L^\prime h_L\|^2\le \|e_{L-1}^K\|^2\). Therefore, \(\|e_L^K\|^2\) is monotonically decreasing and convergent and we have
\begin{equation}
\begin{split}
\label{deqn_ex1a}
\|e_L^K\|^2&-\left(r+\mu_L\right)\|e_{L-1}^K\|^2\\&=\|e_{L-1}^K-\phi_L\left(h_L,X\right)\beta_L^K\|^2-\left(r+\mu_L\right)\|e_{L-1}^K\|^2\\&=\|e_{L-1}^K-\beta_L^lh_L-\phi_L^\prime\left(h_L,X\right)\beta_L^{K\prime}\|^2-\left(r+\mu_L\right)\|e_{L-1}^K\|^2\\&\le\|e_{L-1}^K-\beta_L^\ast h_L\|^2-\left(r+\mu_L\right)\|e_{L-1}^K\|^2\\&\le\|e_{L-1}^K-\beta_L^\prime h_L\|^2-\left(r+\mu_L\right)\|e_{L-1}^K\|^2\\&=\sum_{q=1}^{M}\langle e_{L-1,q}^K-\beta_{L,q}^\prime h_L,e_{L-1,q}^K-\beta_{L,q}^\prime h_L\rangle \\&-\sum_{q=1}^{M}\left(r+\mu_L\right)\langle e_{L-1,q}^K,e_{L-1,q}^K\rangle \\&=\sum_{q=1}^{M}(1-r-\mu_L)\langle e_{L-1,q}^K,e_{L-1,q}^K\rangle\\&-\sum_{q=1}^{M}2\langle e_{L-1,q}^K,\beta_{L,q}^\prime h_L\rangle +\sum_{q=1}^{M}\langle \beta_{L,q}^\prime h_L,\beta_{L,q}^\prime h_L\rangle \\&=\left(1-r-\mu_L\right)\|e_{L-1}^K\|^2-\frac{\sum_{q=1}^{M}\langle e_{L-1,q}^K,h_L\rangle^2}{\|h_L\|^2}\\&=\delta_L^K-\frac{\sum_{q=1}^{M}\langle e_{L-1,q}^K,h_L\rangle^2}{\|h_L\|^2}\\&\le \delta_L^K-\frac{\sum_{q=1}^{M}\langle e_{L-1,q}^K,h_L\rangle^2}{b_n^2}<0.
\end{split}
\end{equation}
which implies \(\underset{L\rightarrow+\infty}\lim{\|e_L^K\|=0}\) and this completes the proof of Theorem 1.

Theorem 1 indicates that the proposed KSCNs hold the universal approximation property. Different from conventional SCNs, the calculation scheme of output weights is conducted on a high-dimensional feature space. This property makes KSCNs more suitable for dealing with strong nonlinearities, such as the real-world industrial cases.
\subsection{Training Strategy of KSCNs}
The main goal of training a neural network is to obtain a network model with both good learning and generalization performances. If a neural network is overtraining, which means this network performs better and better over the training dataset, the overfitting situation is expected to emerge. The early stopping strategy is an effective approach to avoid this situation and is commonly used for training the neural networks [24]. In this work, a training strategy based on early stopping method is designed for KSCNs to obtain the optimal architecture with satisfied generalization performance. It is noted that the early stopping for KSCNs means the termination of training before the model reaches its predefined maximum hidden nodes \(L_{max}\). This strategy not only is helpful for avoiding the overfitting problem, but also can save substantial computational costs.

For a collection of historical data \([X,Y]\), a portion of them is used as the training dataset for model building and the others are used as the validation dataset for assessing generalization performance, which are denoted as \([X_{tr},Y_{tr}]\) and \([X_{tv},Y_{tv}]\), respectively. Predefine the maximum tolerance value \(p_{max}\) and a patience counter with the initial value of 0, which is \(p=0\). Assuming a KSCN model with \(L-1\) (\(L=1,2,...\)) hidden nodes has been established, which is denoted as \(f_{L-1}^K(X_{tr})\). The current validation error can be formulated as \(e_v^{L-1}=Y_{tv}-f_{L-1}^K(X_{tv})\). For the to-be-configured \(L\)th hidden node, the training process continues if \(\|e_v^{L-1}\|^2>\|e_v^{L}\|^2\). Otherwise, set \(p=p+1\). The training process stops until \({p\geq p}_{max}\). In this way, the training process can be terminated in a timely manner to avoid the overfitting situation. Furthermore, the computational cost is also relieved since the final structure of KSCNs may have substantially less hidden nodes than that with the predefined maximum hidden nodes \(L_{max}\).

\subsection{Algorithm Description}

\begin{figure}
    \centering
    \includegraphics[width=1\linewidth]{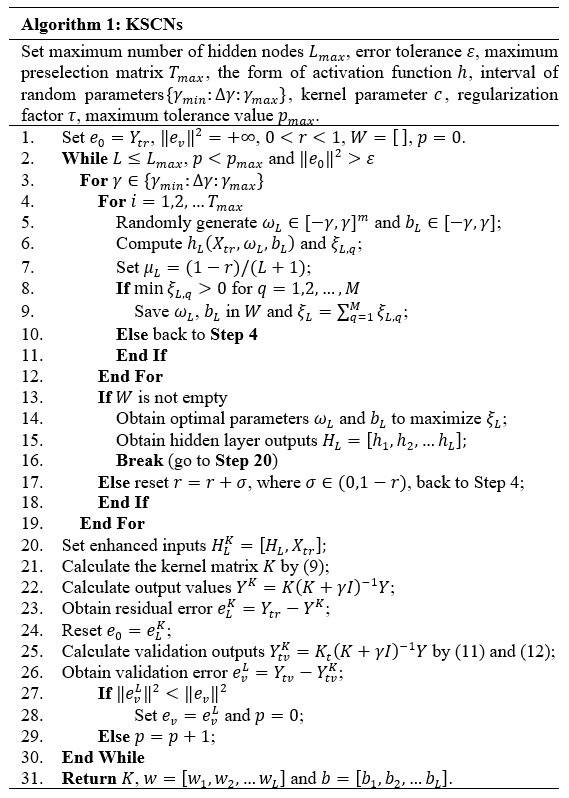}
    %\caption{Enter Caption}
    %\label{fig:enter-label}
\end{figure}

Given training data \(X=\left[x_1,x_2,\ldots,\ x_n\right]^T\in\mathbb{R}^{n\times m}\), \(Y=\left[y_1,y_2,\ldots,y_n\right]^T\in\mathbb{R}^{n\times M}\) and divide them into training dataset \([X_{tr},Y_{tr}]\) and validation dataset \([X_{tv},Y_{tv}]\). Denoting \(e_{L-1}^K\left(X\right)=[e_{L-1,1}^K\left(X\right),e_{L-1,2}^K\left(X\right),\ldots,e_{L-1,M}^K\left(X\right)]\in \mathbb{R}^{n\times M}\) as the residual error matrix with \(L-1\) hidden nodes, where \(e_{L-1,q}^K\left(X\right)=[e_{L-1,q}^K\left(x_1\right),e_{L-1,q}^K\left(x_2\right),\ldots,e_{L-1,q}^K\left(x_n\right)]^T\in\mathbb{R}^n\) for \(q=1,2,\ldots,M\). Define variable
\begin{equation}
\begin{split}
\label{deqn_ex1a}
\xi_{L,q}&=\frac{({e_{L-1,q}^K\left(X\right)}^T\cdot h_L\left(X,\omega_L,b_L\right))^2}{h_L^T\left(X,,\omega_L,b_L\right)\cdot h_L\left(X,\omega_L,b_L\right)}\\&-\left(1-r-\mu_L\right){e_{L-1,q}^K\left(X\right)}^Te_{L-1,q}^K\left(X\right)
\end{split}
\end{equation}
which is used for selecting the most suitable parameters in the algorithm. The detailed procedures are demonstrated in Algorithm 1.
\subsection{Performance Analysis}
\begin{figure*}[!t]
\centering
\includegraphics[width=5 in]{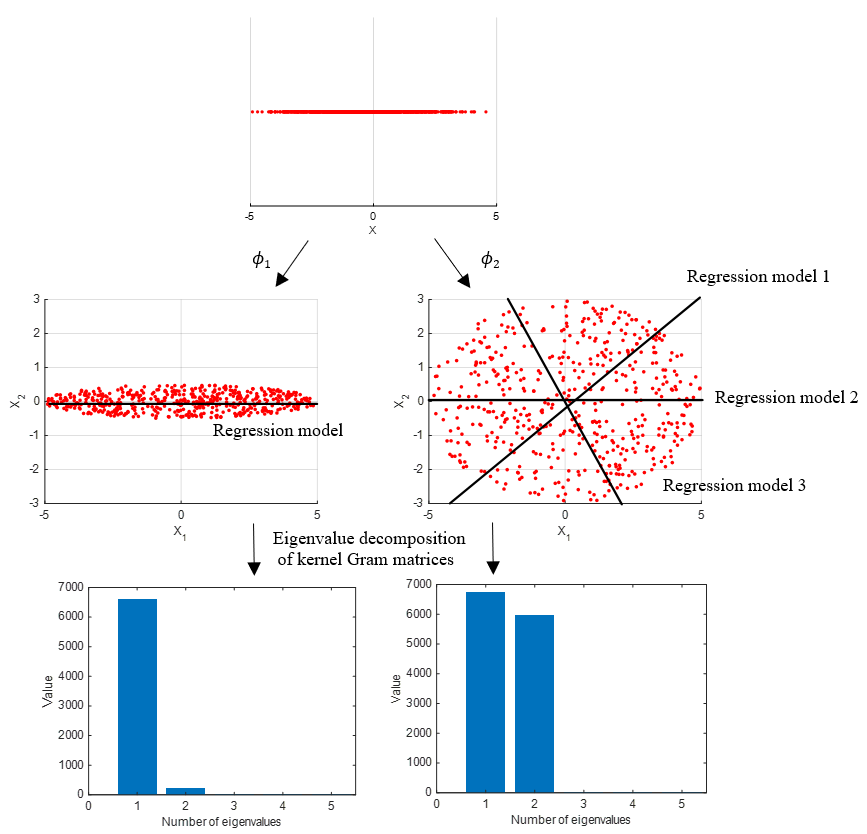}
\caption{An illustration about how different distributions of Gram matrices’ eigenvalues affect the regression performances.}
\label{fig_1}
\end{figure*}
\begin{figure*}[!t]
\centering
\includegraphics[width=4 in]{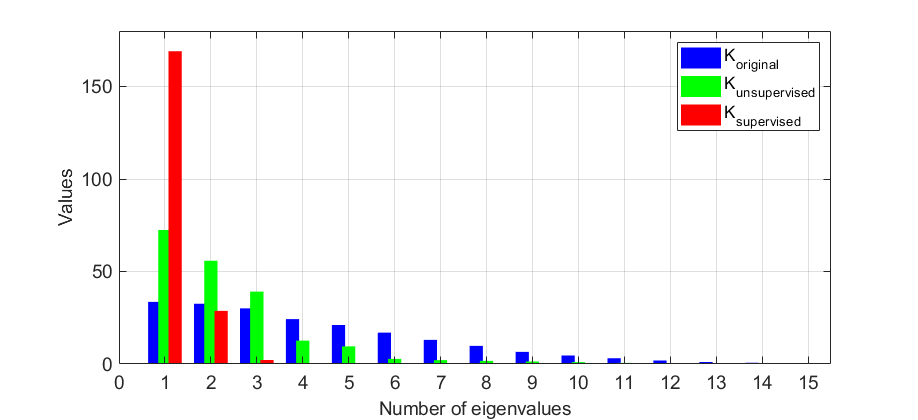}
\caption{The distributions of eigenvalues of different kernel Gram matrices.}
\label{fig_1}
\end{figure*}
The unique feature of the proposed KSCNs is the construction of the high-dimensional feature space, which is based on the supervisory mechanism of SCNs framework. This operation is observed to change the data distribution in high-dimensional feature space.

Since the specific form of nonlinear mapping \(\phi\) is unknown, we focus on the kernel Gram matrix of a KSCN model with \(L\) hidden nodes:

\begin{equation}
K=
\left[
\begin{array}{ccc}  % 三行就写ccc
\phi [f_L(1)]\phi[f_L(1)]^T & \cdots & \phi[f_L(1) ]\phi[f_L(n)]^T \\
\vdots & \ddots & \vdots \\
\phi[f_L(n) ] \phi[f_L(1) ]^T&   \cdots & \phi[f_L(n)]\phi[f_L(n)]^T
\end{array}
\right ]                               % matrix1  
\end{equation}
where \(f_L(i)=H_L (i),x_i^T\), \(i=1,2,...n\). This matrix is used to investigate data characteristics in high-dimensional feature space. According to the theory of kernel principal component analysis (KPCA) [17], [25], the kernel Gram matrix reflects the variance distribution of mapped data in high-dimensional space, which is expressed by the eigenvalues of the matrix. It is desirable that, in a regression task, the distribution of a Gram matrix’s eigenvalues is concentrated so that it is easier to build a regression model for describing the data pattern. An illustration about how the eigenvalue distribution of a Gram matrix influences the regression performance is shown in Fig. 4. Assuming a collection of one-dimensional data samples \(X\), two types of high-dimensional mappings \(\phi_1\) and \(\phi_2\) are used to project the original data onto two-dimensional spaces. The data distribution in the high-dimensional space created by \(\phi_1\) is highly concentrated, which means there are few major eigenvalues in the eigenvalue decomposition process. In this situation, it is easier to establish a regression model for describing data patterns with high modeling accuracy. While the data distribution in another case is more decentralized, therefore, it is difficult to find an accurate regression model to fit this data distribution since there is a great number of potential models can be constructed with a similar (poor) level of modeling performance. Thus, a concentrated distribution of eigenvalues is favorable for building a regression model to describe the data pattern appropriately.

With the analysis above, we now investigate the eigenvalue decomposition of kernel Gram matrix of KSCNs. A function \(f(x)=0.2e^{-(10x-4)^2 }+0.5e^{-(80x-40)^2 }+0.3e^{-(80x-20)^2 }\) [26] is used to study the properties of KSCNs’ kernel Gram matrix. Fig. 5 demonstrates the eigenvalue distributions of three Gram matrices in a descending order, which are: 1) the original kernel \(K_{i,j}=k(x_i,x_j)\) which is termed \(K_{original}\). 2) the kernel with random bases \(K_{i,j}=k(h_i,h_j)\) where \(h_i=[g(\omega _i^T x_1+b_i ),…,g(\omega_i^T x_n+b_i )]^T\) with thoroughly random assignments for weights and biases, and it is termed \(K_{unsupervised}\). 3) the kernel with supervised random bases \(K_{i,j}=k(h_i,h_j)\) where \(h_i=[g(\omega_i^T x_1+b_i ),…,g(\omega_i^T x_n+b_i )]^T\) under SCNs’ supervisory framework for weights and biases, and it is termed \(K_{supervised}\). From the results of Fig. 5, it is clear that the original kernel Gram matrix has a decentralized distribution, which means it is hard for a regression model to capture the principal components information among data. This situation may lead to a difficult condition for regression building, as the illustration shown in Fig. 4. Compared to original kernel, using unsupervised random bases improves the data distribution in feature space, which means the nonlinear transformation by activation functions is helpful for reconstructing data distribution in the feature space. The supervised kernel Gram matrix has the most centralized distribution of eigenvalues, which indicates the data distribution in high-dimensional space has one major direction. This property is desirable for establishing an accurate regression model for describing data patterns.

\section{Performance Evaluation}
In this section, a typical numerical example is used first to demonstrate the effectiveness of our proposed KSCNs. Then, two real-world industrial cases are utilized to evaluate the prediction performances of KSCNs. For comprehensive performance evaluation, the proposed KSCNs are compared with some representative nonlinear regression models besides SCNs, which are RVFL, MLP, SVR and RBFN. SVR and RBFN are two typical kernel-based leaner models, and MLP and RVFL are the representative leaner models with network structure. In this work, the hyperbolic tangent sigmoid function is used as the activation function for all network-based learner models, and the Gaussian kernel function is adopted for all kernel-based learner models. The parameters of all models are assigned by searching for the most suitable parameter combinations which provide the best validation performances. The details are as follows: 

1) RVFL: the search ranges of the number of hidden nodes and the random parameter interval are [1: 1: 100] and ±[0.5, 1, 5, 10, 30, 50, 100, 150, 200, 250], respectively.

2) SVR: the search range of Gaussian kernel parameter is \([10^{-2}: 0.1: 10^2]\). The values of penalty factor and Epsilon are automatically selected and fine-tuned by the MATLAB SVM toolbox.

3) RBFN: the search ranges of Gaussian kernel parameter and the number of centers are \([10^{-2}: 0.1: 10^2]\) and [1: 1: 100], respectively. The centers are randomly selected from training data.

4) MLP: the search ranges of the number of hidden nodes and learning rate are [1: 1: 100] and [0.1, 0.01, 0.001, 0.0001]. The initial weights and biases are set based on the Xavier initialization. BP algorithm is used to fine tune model parameters with a batch size of 1, and the early stopping strategy is used in the iteration procedures with a maximum tolerance value (patience) of 5.

5) SCNs: the random parameter interval is \(\gamma\in[0.5,1,5,10,30,50,100,150,200,250]\) and the number of candidate nodes is set 50. The number of hidden nodes is determined by early stopping strategy with a maximum tolerance value (patience) of 5.

6)	KSCNs: the search ranges of Gaussian kernel parameter and regularization factor are \([10^{-2}: 0.1: 10^2]\) and [0.1, 0.01, 0.001, 0.0001], respectively. The rest of parameters are set the same as those of SCN.

To quantify the modeling performance of different models, two commonly used indicators are adopted in this work, which are the root mean squared error (RMSE) and coefficient of determination (R\(^2\)):
\begin{equation}
\begin{split}
\label{deqn_ex1a}
\mathrm{RMSE}=\sqrt{\frac{\sum\left(Y_t-\widehat{Y_t}\right)^2}{N_t}}
\end{split}
\end{equation}
\begin{equation}
\begin{split}
\label{deqn_ex1a}
\mathrm{R}^\mathrm{2}=1-\frac{\sum\left(Y_t-\widehat{Y_t}\right)^2}{\sum\left(Y_t-\bar{Y_t}\right)^2}
\end{split}
\end{equation}
where \(Y_t\) are the real values of testing data, \(\widehat{Y_t}\) represent the predictive values produced by a learner model, and \(N_t\) is the number of testing samples. RMSE reflects the general error of prediction results, while R\(^2\) demonstrates the ability of explaining the total variance in the quality data.
\subsection{Numerical Example}
\begin{figure}[!t]
\centering
\includegraphics[width=2.5 in]{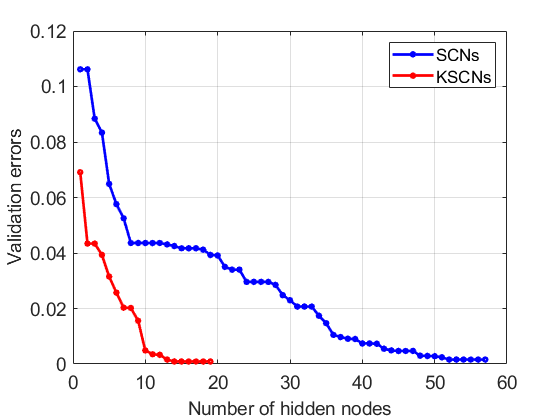}
\caption{Validation processes with early stopping for configuring the hidden node quantities of SCNs and KSCNs.}
\label{fig_1}
\end{figure}
\begin{figure}[!t]
\centering
\includegraphics[width=2.5 in]{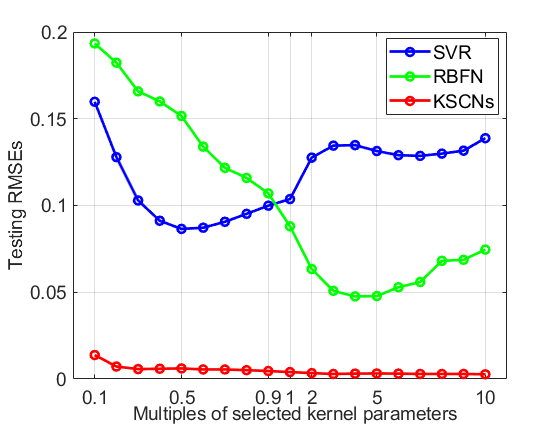}
\caption{Different models’ robustness tests with respect to kernel parameter settings.}
\label{fig_1}
\end{figure}
\begin{figure}[!t]
\centering
\includegraphics[width=2.5 in]{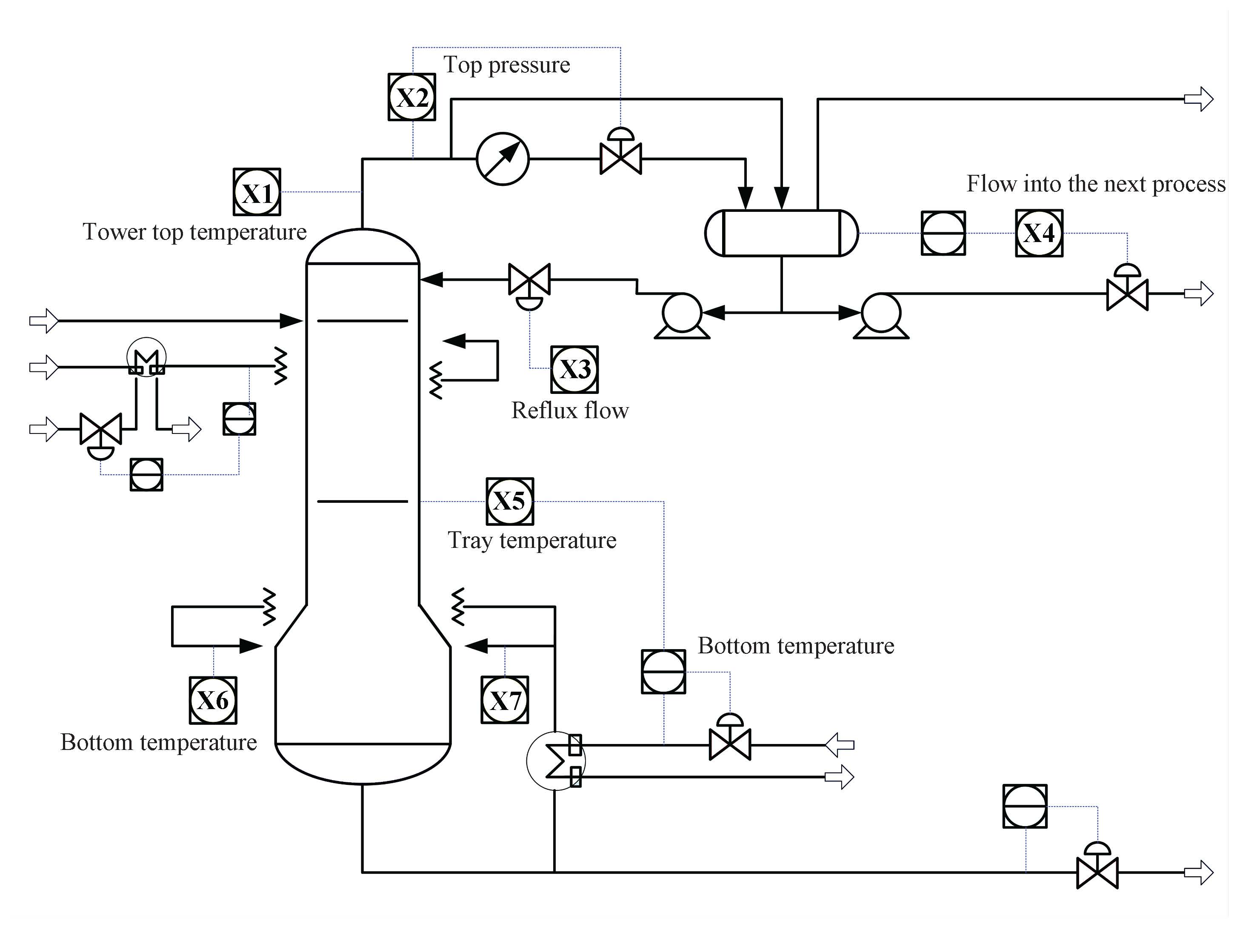}
\caption{Basic flowchart of the debutanizer column.}
\label{fig_1}
\end{figure}

\begin{figure*}[!t]
\centering
\subfloat[]{\includegraphics[width=3in]{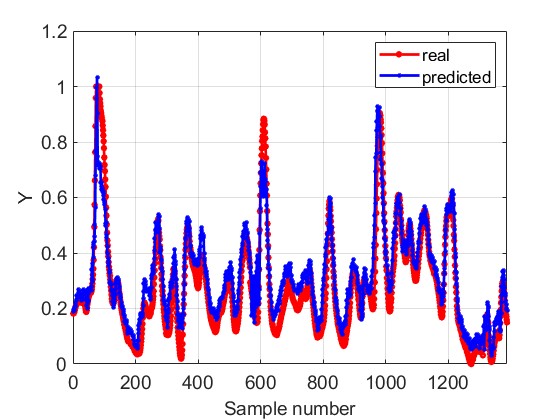}%
\label{fig_first_case}}
\hfil
\subfloat[]{\includegraphics[width=3in]{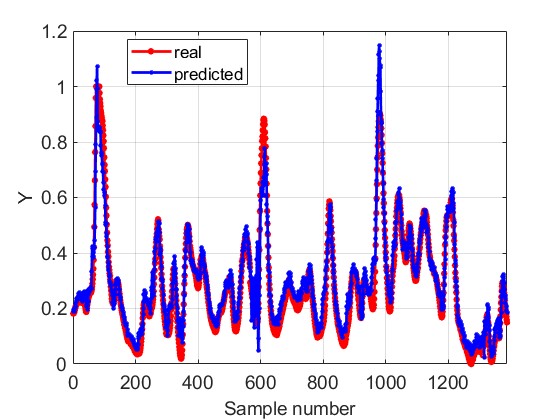}%
\label{fig_second_case}}
\hfil
\subfloat[]{\includegraphics[width=3in]{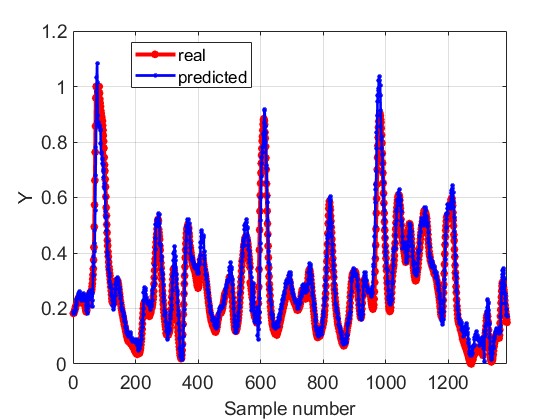}%
\label{fig_third_case}}
\hfil
\subfloat[]{\includegraphics[width=3in]{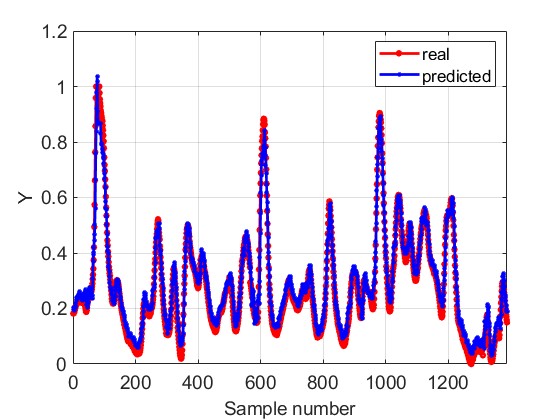}%
\label{fig_forth_case}}
\caption{Detailed prediction results of (a) RVFL, (b) SCNs, (c)KRVFL and (d) KSCNs.}
\label{fig_sim}
\end{figure*}

\begin{table}
    \centering
    \caption{Prediction RMSEs of Different Models (50 Independent Trials)}
    \begin{tabular}{ccccc}
    \hline
    \hline
       Method  & RMSE & R\(^2\) & Min RMSE & Max RMSE\\
       \hline
        MLP & 0.1320±0.0077 & -0.1470±0.3263 & 0.1191 & 0.1518 \\
        
        SVR & 0.1037 & 0.2228 & - & -\\
        
        RBFN & 0.0965±0.0349 & 0.5456±0.4722 & 0.0127 & 0.1729 \\
        
        RVFL & 0.0688±0.0083 & 0.8440±0.0400 & 0.0497 & 0.0866 \\
        
        SCNs & 0.0074±0.0039 & 0.9979±0.0023 & 0.0020 & 0.0206\\
        
        \textbf{KSCNs}&\textbf{0.0032±0.0016}&\textbf{0.9996±0.0006}&\textbf{0.0016}&\textbf{0.0113} \\
        \hline
    \end{tabular}

    \label{tab:my_label}
\end{table}

\begin{table}
    \centering
    \caption{Performance Comparison of KSCNs and SCNs with Different Early Stopping Criteria (50 Independent Trials)}
    \begin{tabular}{c|cc|cc}
    \hline
    \hline
         & \multicolumn{2}{c|}{KSCNs}  & \multicolumn{2}{c}{SCNs}  \\
       \hline
        \(p_{max}\) & RMSE & Nodes & RMSE & Nodes\\
        \hline
        1 & 0.0136±0.0209 & \textbf{13.30±3.41} & 0.1527±0.0326 & \textbf{4.66±3.24}\\
        
        3 & 0.0043±0.0020 & 18.60±2.16 & 0.1008±0.0185 & 13.24±7.28 \\
        
        5 & 0.0039±0.0019 & 21.10±2.01 & 0.0501±0.0470 & 44.64±32.10 \\
        
        7 & \textbf{0.0037±0.0017 }& 23.82±2.63 & 0.0195±0.0367 & 70.30±28.06\\
        
        9&0.0038±0.0022&26.10±2.26&\textbf{0.0090±0.0239}&85.16±23.66 \\
        \hline
    \end{tabular}

    \label{tab:my_label}
\end{table}

Assuming input variables \(x\) are generated from the uniform distribution of interval [0,1]. A real-valued function is defined as \(f(x)=0.2e^{-(10x-4)^2 }+0.5e^{-(80x-40)^2 }+0.3e^{-(80x-20)^2 }\) [26]. A total of 600 samples are generated, and half of them is used as the training data for building the regression model (200 for training and 100 for validation). While the other half is used as the testing data for evaluating the prediction performances of different learner models. 

After the training and validation processes, the parameters of different models are selected as follows. For MLP, the number of hidden nodes is 33, and the learning rate is 0.0001. For SVR, the Gaussian kernel parameter is 0.11. For RBFN, the number of centers is 43 and the Gaussian kernel parameter is 0.01. For RVFL, the number of hidden nodes is 98 and the random parameter interval is [-10,10]. For SCNs, the number of hidden nodes is 57. For KSCNs, the number of hidden nodes is 19, the Gaussian kernel parameter is 1.21 and the regularization factor is 0.0001.

Fig. 6 demonstrates the validation losses of SCNs and KSCNs. The validation loss of KSCNs is generally smaller than that of SCNs and the convergence rate is also much faster. With a maximum tolerance value (patience) of 5, the training process of KSCNs terminates at the 19th node, while the same situation occurs for SCNs at the 57th node. The results demonstrate that, with the assistance of kernel technique, KSCNs have stronger representation learning capability in nonlinear regression cases compared to SCNs, so that a smaller number of nodes of KSCNs can describe the same nonlinear expression.

Table I gives the quantified performances of these methods, and the results of all methods were obtained from 50 times independent trials. Clearly, the proposed KSCNs demonstrate the best prediction performance by providing the minimum comprehensive RMSEs and the maximum R\(^2\). Furthermore, as a randomized learner model, KSCNs provide more stable performances than SCNs, which is illustrated by producing a smaller value of standard deviation of RMSEs under 50 trials. It is indicated that the proposed KSCNs are less affected by different forms of initial random parameter setting. Performance stability is an important indicator for a randomized leaner model, which demonstrates the reliability of a learner model in real-world applications.

The early stopping strategy is significant for preventing the neural networks from overfitting. The results from Fig. 6 also verify the effectiveness of early stopping. To comprehensively compare the influences of early stopping strategy for SCNs and KSCNs, Table II gives the prediction performances and structure details of these two models under different stopping criteria, which is represented by the predefined values of maximum tolerance (\(p_{max}\)). With the results of Table II, the following conclusions can be drawn: 1) the performances of KSCNs are less sensitive to different stopping criteria. 2) the structures of KSCNs under different stopping criteria change slightly, which is favorable for performance stability. 3) the model structures with different stopping criteria of KSCNs maintain at a similar level. 4) different stopping criteria have huge impacts on SCNs’ prediction performances and model structures. In conclusion, the representation learning capability of the network is enhanced by introducing the kernel technique, resulting in a fast convergency rate in the training and validation processes.

Parameter sensitivity is another significant indicator for learner models, especially for those kernel-based leaner models. A simple and common approach to assign kernel parameters is to select a parameter which drives the leaner model to perform well over validation data. However, the distribution of validation data is not usually the same as that of testing data, which means the selected parameter may not guarantee good generalization performances. A reliable learner model is expected to maintain good performances over a certain range of model parameters. Fig. 7 illustrates the kernel parameter sensitivity tests of three kernel based leaner models, which shows the prediction RMSEs of different models over a range from 0.1 times to 10 times selected parameter (the RMSE of each approach is taken average under 50 independent trials). In the figure, KSCNs provide a stable performance over this wide range of kernel parameters, while the performances of SVR and RBFN are sensitive to the kernel parameter settings. The reason for that is due to the different kernel mapping operations of these three methods. In SVR and RBFN, the original raw data are used for high-dimensional feature space construction. Differently, KSCNs use the random bases, which are generated by the supervisory mechanism of SCNs, along with the raw data for constructing a high-dimensional feature space. Therefore, the performances of KSCNs are verified to be less sensitive to the kernel parameter settings.

\subsection{Debutanizer Column Application}

\begin{figure}[!t]
\centering
\includegraphics[width=2.5 in]{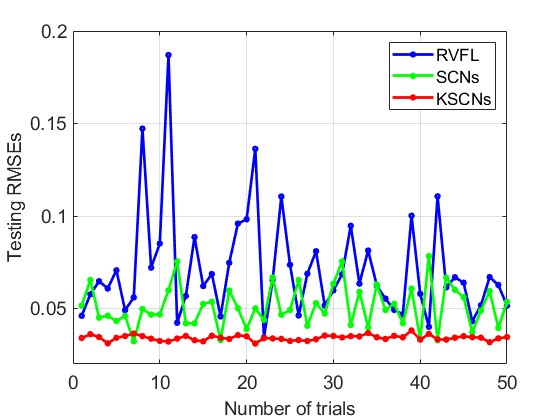}
\caption{Performance stability tests of different randomized learner models.}
\label{fig_1}
\end{figure}

\begin{figure}[!t]
\centering
\includegraphics[width=2.5 in]{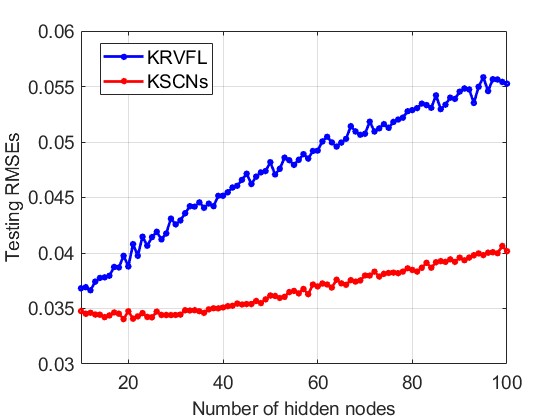}
\caption{Performance comparison of KRVFL and KSCNs under different structures.}
\label{fig_1}
\end{figure}

\begin{figure}[!t]
\centering
\includegraphics[width=2.5 in]{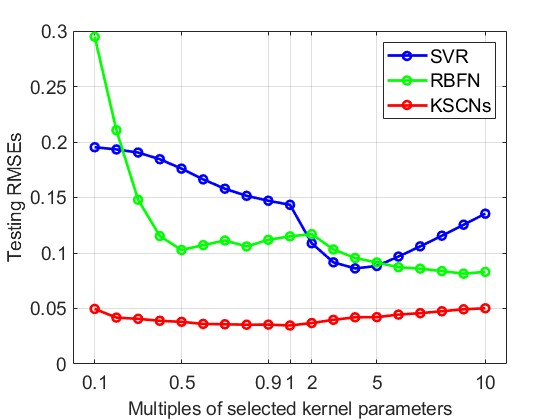}
\caption{Different models’ robustness tests with respect to kernel parameter settings.}
\label{fig_1}
\end{figure}

\begin{table}
    \centering
    \caption{Prediction RMSEs of Different Models (50 Independent Trials)}
    \begin{tabular}{ccccc}
    \hline
    \hline
       Method  & RMSE & R\(^2\) & Min RMSE & Max RMSE\\
       \hline
        MLP & 0.1534±0.0171 & -1.1037±0.9516 & 0.1272 & 0.1959 \\
        
        SVR & 0.1434 & -2.8950 & - & -\\
        
        RBFN & 0.1140±0.0240 & 0.6087±0.1761 & 0.0708 & 0.1947 \\
        
        RVFL & 0.0711±0.0287 & 0.8674±0.0880 & 0.0343 & 0.1869 \\
        
        SCNs & 0.0511±0.0114 & 0.9168±0.0433 & 0.0322 & 0.0783\\
        
        \textbf{KSCNs}&\textbf{0.0342±0.0014}&\textbf{0.9625±0.0040}&\textbf{0.0310}&\textbf{0.0381} \\
        \hline
    \end{tabular}

    \label{tab:my_label}
\end{table}

\begin{table}
    \centering
    \caption{Performance Comparison of KSCNs and SCNs with Different Early Stopping Criteria (50 Independent Trials)}
    \begin{tabular}{c|cc|cc}
    \hline
    \hline
         & \multicolumn{2}{c|}{KSCNs}  & \multicolumn{2}{c}{SCNs}  \\
       \hline
        \(p_{max}\) & RMSE & Nodes & RMSE & Nodes\\
        \hline
        1 & 0.0370±0.0021 & \textbf{3.38±1.43} & 0.0970±0.0293 & \textbf{3.70±1.42}\\
        
        3 & 0.0352±0.0019 & 11.34±5.69 & 0.0618±0.0292 & 10.46±5.64 \\
        
        5 & \textbf{0.0348±0.0018} & 26.72±18.06 & \textbf{0.0558±0.0144} & 18.40±7.33 \\
        
        7 & 0.0363±0.0020& 42.00±23.84 & 0.0658±0.0180 & 24.36±4.53\\
        
        9&0.0374±0.0020&61.44±30.22&0.0759±0.0195&28.18±4.42 \\
        \hline
    \end{tabular}

    \label{tab:my_label}
\end{table}

In the petroleum refining industry, there is an important unit used to split naphtha and desulfuration, which is called the debutanizer column [27]. In the process, there are six devices located at different positions, which are heat exchanger, overhead condenser, head reflux pump, reflux accumulator, bottom reboiler, and feed pump to the LPG splitter. The debutanizer column used in this paper is in Syracuse, Italy. The basic schematic of the process can be found in Fig. 8.

In this process, propane (C3) and butane (C4) are required to be removed. For that purpose, the content of butane in the bottom product is required to be known. Unfortunately, the butane content is usually measured by a gas chromatograph which is installed in the overhead of the column, and this gas chromatograph is far from the process. This situation greatly affects the awareness of the butane content, and it is harmful to the quality control of the process. Therefore, it is of great necessity to establish the regression model of the butane content for producing the content values in time, which is also called a soft sensor in industrial processes [28]-[30]. Based on the prior knowledge of the process, seven routine-measured variables are selected as the predictors for the vital variables, which are \(X_1\): top temperature, \(X_2\): top pressure, \(X_3\): reflux flow, \(X_4\): flow to next process, \(X_5\): 6th tray temperature, \(X_6\): bottom temperature A and \(X_7\): bottom temperature B. Based on the physiochemical insight and expert knowledge, an optimal augmentation strategy is employed for better describing the relation between predictors and vital variables, which is formulated as [27]
\begin{equation}
\begin{split}
\label{deqn_ex1a}
y\left(k\right)=&f(X_\mathrm{1}\left(k\right),X_\mathrm{2}\left(k\right),...,X_\mathrm{5}\left(k\right),X_\mathrm{5}\left(k-1\right),\\&X_\mathrm{5}\left(k-\mathrm{2}\right),X_\mathrm{5}\left(k-\mathrm{3}\right),\frac{X_\mathrm{6}\left(k\right)+X_\mathrm{7}\left(k\right)}{\mathrm{2}},\\&y\left(k-\mathrm{4}\right),y\left(k-\mathrm{5}\right),y\left(k-\mathrm{6}\right)).
\end{split}
\end{equation}

A total of 2394 samples are collected from historical data. Among them, the first 1000 samples are used as training data (800 for training and 200 for validation) and the rest are used as testing data for evaluating prediction performances. After the training and validation processes, the parameters of different models are assigned as follows. For MLP, the number of hidden nodes is 56, and the learning rate is 0.01. For SVR, the Gaussian kernel parameter is 8.11. For RBFN, the number of centers is 76 and the Gaussian kernel parameter is 1.91. For RVFL, the number of hidden nodes is 17 and the random parameter interval is [-1,1]. For SCNs, the number of hidden nodes is 19. For KSCNs, the number of hidden nodes is 12, the Gaussian kernel parameter is 99.91 and the regularization factor is 0.001.

Table III gives the prediction performances over testing data of different learner models and Fig. 9 shows the detailed prediction results of some models. From Table III, it is clear that our proposed KSCNs obtain the best generalization performance in this case. Different from the numerical case, the industrial cases are with unclear distribution and dynamic information. Hence, it is easy for the learner models to fall into the local optimum situation (such as MLP in this case). Table IV also gives the performance comparison of SCNs and KSCNs under different early stopping criteria. The performances under different values of patience of KSCNs maintain consistent, while the performances of SCNs fluctuates with different situations. It is noted that the numbers of nodes of SCNs under different stopping criteria are generally smaller than those of KSCNs, which means the training processes of SCNs are with greater chance to be terminated at an early stage in this case. That indicates more hidden nodes of SCNs are not always helpful for reducing the validation errors. For KSCNs, the model architecture scales become larger with the raising values of patience. This result verifies the effectiveness of KSCNs model for continuously improving the validation performance with the number of hidden nodes increasing. 

In addition to prediction performance, the proposed KSCNs also provide the best performance stability among these random learner models. Fig. 10 demonstrates the prediction RMSEs of RVFL, SCNs and KSCNs. RVFL provide the most unstable outputs due to the thoroughly random configurations of input weights and biases. With a well-designed supervisory mechanism, SCNs can generate quality-relevant hidden nodes which not only reduce the prediction errors but also improve the performance stability. By introducing the kernel techniques, KSCNs greatly deduct the strong nonlinearities among industrial data, resulting in the easy construction of regression model in feature space. Therefore, the prediction performances of KSCNs maintain at a similar level under 50 independent trails.

To verify the effectiveness of supervisory mechanism of KSCNs, Fig. 11 demonstrates the performance comparison (the RMSE of each approach is taken average under 50 independent trials) of KSCNs and kernel RVFL (KRVFL) [31], [32] whose model parameters are assigned under the same principles as those of KSCNs. KRVFL can be considered as a kernel-based randomized leaner model, where the modeling process is similar to the proposed KSCNs except for the supervisory mechanism when assigning weights and biases. Compared to RVFL and SCNs, the generalization performance of KRVFL is improved due to the kernel mapping technology. Compared to KSCNs, the prediction performance of KRVFL under different number of nodes is worse and unstable. The reason for this result is that the random bases of KRVFL are generated in an unrestricted way, which makes the modeling accuracy greatly depends on the forms of random inputs and kernel parameters. For KSCNs, the random bases are data-dependent due to the supervisory mechanism in the modeling process. With the number of hidden nodes growing at the early stage, the hidden layer of KSCNs is becoming more quality-relevant and the prediction RMSE is accordingly declines. When the number of hidden nodes continuously increases, the overfitting situation occurs so that the prediction RMSE starts to rise. For KRVFL, due to the completely random mechanism for generating hidden nodes, the prediction RMSE keeps growing with the number of hidden nodes increasing. The result verifies the effectiveness of the supervisory mechanism of KSCNs for improving generalization performance of learner models.

Kernel parameter selection is significant for kernel-based learner models. The way for selecting kernel parameters in this paper is based on the performances over validation dataset, which is effective for avoiding the overfitting situation. However, if the learner models are sensitive to kernel parameter settings, the selected parameter may not be the most suitable one for generalization performance due to the differences between validation and testing data. Fig. 12 demonstrates kernel parameter sensitivity tests of SVR, RBFN and KSCNs, where the prediction RMSEs accordingly change with kernel parameter changing from 0.1 times selected parameter to 10 times selected parameter (the prediction RMSE is taken average value under 50 independent trials). It can be observed that the curve of KSCNs is much smoother than those of SVR and RBFN, which illustrates the proposed model is less sensitive to the kernel parameters. The reason for this result is obvious: the kernel matrices of SVR and RBFN are based on the original input data, while KSCNs utilize the random bases which are generated under the supervisory mechanism for kernel mapping. Therefore, KSCNs can provide more stable performances over a wide range of kernel parameters.

\subsection{Power Load Forecasting}

\begin{figure}[!t]
\centering
\includegraphics[width=2.5 in]{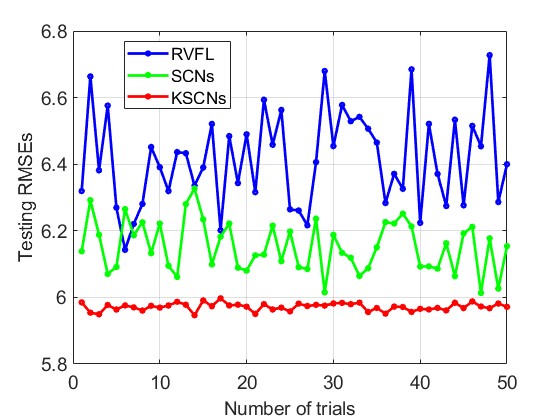}
\caption{Performance stability tests of different randomized learner models.}
\label{fig_1}
\end{figure}

\begin{figure}[!t]
\centering
\includegraphics[width=2.5 in]{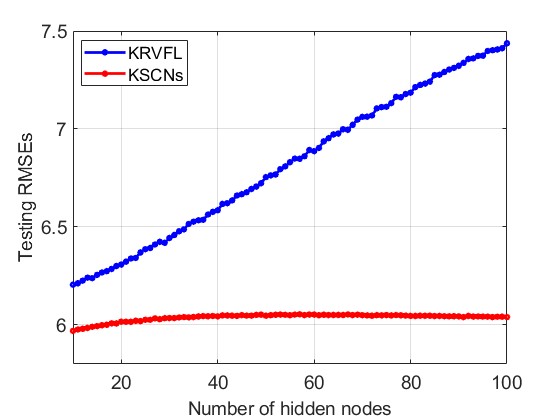}
\caption{Performance comparison of KRVFL and KSCNs under different structures.}
\label{fig_1}
\end{figure}

\begin{table}
    \centering
    \caption{Prediction RMSEs of Different Models (50 Independent Trials)}
    \begin{tabular}{ccccc}
    \hline
    \hline
       Method  & RMSE & R\(^2\) & Min RMSE & Max RMSE\\
       \hline
        SVR & 7.6097 & -0.0063 & - & - \\
        
        RBFN & 7.4100±1.5947 & 0.5644±0.2325 & 6.0162 & 12.4322\\
        
        MLP & 7.0304±0.4402 & 0.6891±0.0335 & 6.2042 & 8.0070 \\
        
        RVFL & 6.4147±0.1399 & 0.7214±0.0118 & 6.1428 & 6.7274 \\
        
        SCNs & 6.1523±0.0757 & 0.7482±0.0062 & 6.0132 & 6.3270\\
        
        \textbf{KSCNs}&\textbf{5.9716±0.0114}&\textbf{0.7529±0.0016}&\textbf{5.9464}&\textbf{5.9972} \\
        \hline
    \end{tabular}

    \label{tab:my_label}
\end{table}

\begin{table}
    \centering
    \caption{Performance Comparison of KSCNs and SCNs with Different Early Stopping Criteria (50 Independent Trials)}
    \begin{tabular}{c|cc|cc}
    \hline
    \hline
         & \multicolumn{2}{c|}{KSCNs}  & \multicolumn{2}{c}{SCNs}  \\
       \hline
        \(p_{max}\) & RMSE & Nodes & RMSE & Nodes\\
        \hline
        1 & \textbf{5.9201±0.0170} & \textbf{3.18±1.29} & 6.8401±1.2421 & \textbf{4.22±1.45}\\
        
        3 & 5.9663±0.0256 & 9.62±4.13 & 6.1858±0.2081& 8.24±3.23 \\
        
        5 & 5.9832±0.0286 & 14.54±7.05 & \textbf{6.1278±0.1436} & 12.56±3.86 \\
        
        7 & 6.0031±0.0258& 19.78±8.62 & 6.1312±0.0893 & 14.24±4.57\\
        
        9&6.0240±0.0232 & 27.02±10.60&6.1364±0.1103&18.42±6.41 \\
        \hline
    \end{tabular}

    \label{tab:my_label}
\end{table}

Power load forecasting is important for the operation and production planning in modern power systems. And the stable operation of power systems has a crucial influence on the development of national economy. However, accurate power load forecasting is difficult for the power grid due to its strong nonlinear, nonstationary, seasonal and some unpredictable reasons. Therefore, developing an applicable prediction model for producing the reliable forecasting of power load is of great significance. This application uses the load data collected from a specific 500kV substation in Liaoning, China. The load data was collected once an hour from January to February 2023, which indicates a total of 1417 samples are collected. According to prior knowledge, four routinely measured variables are expected to have a great impact on the short-term electricity demand (\(y\)) prediction, which are \(X_1\): temperature, \(X_2\): humidity, \(X_3\): precipitation and \(X_4\): wind speed. Therefore, these four variables are utilized as the predictors of the power load forecasting. Take the dynamic information into consideration, an augmentation formulation is used to build the prediction model, which is \(y\left(k\right)=f\left(X_\mathrm{1}\left(k\right),X_\mathrm{2}\left(k\right),X_\mathrm{3}\left(k\right),X_\mathrm{4}\left(k\right),y\left(k-1\right)\right)\). The first half collected data is used as training data and the other half is for testing. Gaussian noise is employed to the testing data to form the validation data.

According to the results of training and validation processes, the parameters for different models are set as follows. For MLP, the number of hidden nodes is 15, and the learning rate is 0.01. For SVR, the kernel parameter is 49.71. For RBFN, the number of centers is 5 and the kernel parameter is 4.81. For RVFL, the number of hidden nodes is 94 and the random parameter interval is [-5,5]. For SCNs, the number of hidden nodes is 22. For KSCNs, the number of hidden nodes is 10, the kernel parameter is 98.61 and the regularization factor is 0.1.

Table V shows the quantified prediction performances and Fig. 13 shows the detailed prediction RMSEs of three randomized methods under 50 independent trials. The proposed KSCNs outperform other models by providing the most favorable performance. Furthermore, KSCNs exhibit the highest level of performance stability among these randomized methods. For the real-world application, performance stability is significant for maintaining stable operations and avoiding potential risks. To verify the effectiveness of supervisory mechanism of KSCNs, Fig. 14 also illustrates the performance comparison between KSCNs and KRVFL, whose parameters are configured under the same principles. It is clear that the performances of KSCNs maintain a good level with the hidden nodes increasing, while the testing RMSEs of KRVFL rise consistently. Different to KRVFL where the hidden nodes are generated completely randomly, each hidden node of KSCNs is generated under the supervisory mechanism so that the prediction performance can be maintained within a certain level.

Table VI also demonstrates the performance comparison of SCNs and KSCNs under different early stopping criteria. It is noted that the structures of KSCNs generally have a larger scale compared to these with the same stopping criteria of SCNs’. That indicates the training process of KSCNs is more sustainable for continuously declining the validation errors, resulting in better prediction accuracy over testing data.

\subsection{Robustness Analysis to Parameter Settings}

There are two major parameters in the modeling process of KSCNs, which are kernel function parameter and regularization factor. As the experimental results above, the performances of KSCNs under a wide range of kernel parameters hold steady, which means KSCNs are robust to kernel parameter settings. This section focuses on the impacts of different regularization factors on model performances.

The regularization factor is an important parameter used to control model complexity for improving the generalization performance of a learner model [33]. In KSCNs’ modeling procedures, the regularization factor \(\tau\) is introduced in the objective function (7) for good generalization performance to unknown data. The larger the value of regularization factor is, the lower the model complexity is. According to the training strategy of KSCNs, the value of regularization factor is pair-searched along with the kernel parameter based on validation performance. Hence, to observe the influences of regularization factors, we first fix the values of kernel parameters and model structures, then obtain the prediction performances of KSCNs models (shown in Table VII) with the values of regularization factors changing from [0.1, 0.01, 0.001, 0.0001] (DB1 represents the numerical case, DB2 represents the debutanizer column case and DB3 represents the power load forecasting case).

From the results of Table VII, it can be concluded that KSCNs perform differently with different values of regularization factors. That indicates KSCNs are more sensitive to different settings of regularization factors, especially for the case where the training data and the testing data have a similar distribution (such as DB1). This is reasonable due to different levels of model complexities. However, in the training process of KSCNs, the regularization factor and the kernel parameter are assigned simultaneously, which means it is impractical to consider the impact of regularization factor individually. The results of Table VII also show that KSCNs model needs different levels of regularization for different regression tasks, which is mainly attributed to the different patterns of training data. For example, the mathematical expressions of training data and testing data are the same in the numerical case (DB1), so that a small impact of regularization is helpful for obtaining good prediction performance. For industrial cases, however, the training data and testing data may have different distributions, and the prediction RMSEs are not with dramatic changes under different levels of regularizations.

To explore the best generalization abilities under different model complexities, Table VIII shows the prediction RMSEs of KSCNs models with fixed values of regularization factors (the values of kernel parameters are independently searched corresponding to a specific regularization factor). The results here also demonstrate a high level of sensitivities to the settings of regularization factors. That indicates the performance of KSCNs model is easily affected by different levels of model complexities, which is a common situation with the learner models with a regularization strategy.

With the analyses above, some conclusions can be drawn: 1) with the fixed model structure and other parameters, the performance of KSCNs is sensitive to regularization factor settings. 2) to ensure a stable performance of KSCNs model, the value of regularization factor should be pair-searched along with the kernel parameter.

\begin{table}
    \centering
    \caption{Prediction RMSEs with Fixed Kernel Parameters and Corresponding Regularization Factors (50 Independent Trials)}
    \begin{tabular}{c|cccc}
    \hline
    \hline
         \multirow{2}{*}{Data}  & \multicolumn{4}{c}{Regularization Factors} \\ 

         & 0.1 & 0.01 & 0.001 & 0.0001\\
    \hline
        DB1 & 0.1390±0.0015 & 0.0644±0.0281 & 0.0100±0.0037 & \textbf{0.0032±0.0016}\\
       DB2  & 0.0669±0.0027 & 0.0489±0.0020 & \textbf{0.0342±0.0014} & 0.0347±0.0028\\
       DB3  & \textbf{5.9716±0.0114} & 6.0861±0.0297 & 6.1727±0.0691 & 6.5527±0.2073\\
    \hline
    \end{tabular}

    \label{tab:my_label}
\end{table}

\begin{table}
    \centering
    \caption{Prediction RMSEs with Fixed Regularization Factors and Corresponding Kernel Parameters (50 Independent Trials)}
    \begin{tabular}{c|cccc}
    \hline
    \hline
         \multirow{2}{*}{Data}  & \multicolumn{4}{c}{Regularization Factors} \\ 

         & 0.1 & 0.01 & 0.001 & 0.0001\\
    \hline
        DB1 & 0.0719±0.0009 & 0.0447±0.0158 & 0.0085±0.0041 & \textbf{0.0032±0.0016}\\
       DB2  & 0.0654±0.0021 & 0.0451±0.0019 & \textbf{0.0342±0.0014} & 0.0343±0.0026\\
       DB3  & \textbf{5.9716±0.0114} & 6.0714±0.0298 & 6.1513±0.0549 & 6.4843±0.1804\\
    \hline
    \end{tabular}

    \label{tab:my_label}
\end{table}

\section{Conclusion}
Through projecting the enhanced inputs onto a high-dimensional feature space, the random bases of SCNs are used to span a RKHS for regression building. This strategy is verified to improve the data distribution in high-dimensional feature space, where the principal feature information is more distinct to be captured. Under the framework of SCNs, the universal approximation property of our proposed method is also guaranteed. Three case studies are reported to illustrate the effectiveness of our proposed method, including generalization performance, model’s performance stability and robustness with respect to kernel parameter settings. It is worth noting that the proposed method has some deficiencies. For example, two extra parameters are introduced in model building compared to original SCN model, which increases the modeling complexity to some extent. And the computational cost of kernel Gram matrix is quite heavy for large-scale dataset. In the future, fast algorithms for computing kernel matrix will be focused. Also, a lightweight version of KSCNs will be developed by referring to the SCM framework [34], [35]. Moreover, the intrinsic mechanism of how KSCNs improve the data distribution in high-dimensional space will be further investigated.

\vfill


\begin{thebibliography}{1}
\bibliographystyle{IEEEtran}



\bibitem{}
W. Samek, G. Montavon, S. Lapuschkin, C. J. Anders, and K.-R. Muller, “Explaining deep neural networks and beyond: A review of methods and applications,” \textit{Proceedings of the IEEE}, vol. 109, no. 3, pp. 247–278, Mar. 2021.

\bibitem{}
D. E. Rumelhart, G. E. Hinton, and R. J. Williams, “Learning internal representations by back-propagating errors,” \textit{Nature}, vol. 323, no. 6088, pp. 533–536, 1986.

\bibitem{}
H. Lu, R. Setiono, and H. Liu, “Effective data mining using neural networks,” \textit{IEEE Transactions on Knowledge and Data Engineering}, vol. 8, no. 6, pp. 957–961, Dec. 1996.

\bibitem{}
D. Wang, “Editorial: Randomized algorithms for training neural networks,” \textit{Information Sciences}, vols. 364–365, pp. 126–128, Oct. 2016.

\bibitem{}
S. Scardapane and D. Wang, “Randomness in neural networks: An overview,” \textit{Wiley Interdisciplinary Reviews-Data Mining and Knowledge Discovery}, vol. 7, no. 2, pp. 1-18, Apr. 2017.

\bibitem{}
M. Li and D. Wang, “Insights into randomized algorithms for neural networks: Practical issues and common pitfalls,” \textit{Information Sciences}, vols. 382–383, pp. 170–178, Mar. 2017.

\bibitem{}
D. Wang and M. Li, “Stochastic configuration networks: Fundamentals and algorithms,”\textit{ IEEE Transactions on Cybernetics}, vol. 47, no. 10, pp. 3466–3479, Oct. 2017.

\bibitem{}
D. Wang, P. Tian, W. Dai, and G. Yu, “Predicting particle size of copper ore grinding with stochastic configuration networks,” \textit{IEEE Transactions on Industrial Informatics}, vol. 20, no. 11, pp. 12969-12978, Nov. 2024.

\bibitem{}
M. Li and D. Wang, “2-D stochastic configuration networks for image data analytics,” \textit{IEEE Transactions on Cybernetics}, vol. 51, no. 1, pp. 359–372, Jan. 2021.

\bibitem{}
D. Wang and G. Dang, “Recurrent stochastic configuration networks for temporal data analytics,” \textit{arXiv:}2406.16959v2, 2024.

\bibitem{}
D. Wang and G. Dang, “Fuzzy recurrent stochastic configuration networks for industrial data analytics,” \textit{arXiv:} 2407.11038v2, 2024.

\bibitem{}
K. Li, J. Qiao, and D. Wang, “Online self-learning stochastic configuration networks for nonstationary data stream analysis,”\textit{ IEEE Transactions on Industrial Informatics}, vol. 20, no. 3, pp. 3222–3231, Mar. 2024.

\bibitem{}
K. Li, J. Qiao, and D. Wang, “Fuzzy stochastic configuration networks for nonlinear system modeling,” \textit{IEEE Transactions on Fuzzy Systems}, vol. 32, no. 3, pp. 948-957, Mar. 2024.

\bibitem{}
J. Lu and J. Ding, “Mixed-distribution-based robust stochastic configuration networks for prediction interval construction,” \textit{IEEE Transactions on Industrial Informatics}, vol. 16, no. 8, pp. 5099–5109, Aug. 2020.

\bibitem{}
J. Lu, J. Ding, X. Dai, and T. Chai, “Ensemble stochastic configuration networks for estimating prediction intervals: A simultaneous robust training algorithm and its application,” \textit{IEEE Transactions on Neural Networks and Learning Systems}, vol. 31, no. 12, pp. 5426–5440, Dec. 2020.

\bibitem{}
W. Dai, X. Zhou, D. Li, S. Zhu, and X. Wang, “Hybrid parallel stochastic configuration networks for industrial data analytics,”\textit{ IEEE Transactions on Industrial Informatics}, vol. 18, no. 4, pp. 2331-2341, Apr. 2022.

\bibitem{}
B. Schölkopf, A. Smola, and K.-R. Müller, “Nonlinear component analysis as a kernel eigenvalue problem,”\textit{ Neural Computation}, vol. 10, no. 5, pp. 1299–1319, Jul. 1998.

\bibitem{}
B. Schölkopf and A.J. Smola,\textit{ Learning with Kernels}, Cambridge, MA, USA: MIT Press, 2002.

\bibitem{}
K.-R. Müller, S. Mika, G. Rätsch, K. Tsuda, and B. Schölkopf, “An introduction to kernel-based learning algorithms,” \textit{IEEE Transactions on Neural Networks}, vol. 12, no. 2, pp. 181–201, Mar. 2001.

\bibitem{}
M. A. Hearst, S. T. Dumais, E. Osuna, J. Platt, and B. Scholkopf, “Support vector machines,” \textit{IEEE Intelligent Systems and their Applications}, vol. 13, no. 4, pp. 18–28, Jul. 1998.

\bibitem{}
A. J. Smola and B. Schölkopf, “A tutorial on support vector regression,” \textit{Statistics and Computing}, vol. 14, no. 3, pp. 199–222, Aug. 2004.

\bibitem{}
J. Park and I. W. Sandberg, “Universal approximation using radial-basis-function networks,”\textit{ Neural Computation}, vol. 3, no. 2, pp. 246–257, Jun. 1991.

\bibitem{}
Y.-H. Pao and Y. Takefuji, “Functional-link net computing: Theory, system architecture, and functionalities,”\textit{ Computer}, vol. 25, no. 5, pp. 76–79, May 1992.

\bibitem{}
L. Prechelt, “Automatic early stopping using cross validation: Quantifying the criteria,” \textit{Neural Networks}, vol. 11, no. 4, pp. 761–767, Jun. 1998.

\bibitem{}
Y. Motai and H. Yoshida, “Principal composite kernel feature analysis: Data-dependent kernel approach,” \textit{IEEE Transactions on Knowledge and Data Engineering}, vol. 25, no. 8, pp. 1863–1875, Aug. 2013.

\bibitem{}
I. Y. Tyukin and D. V. Prokhorov, “Feasibility of random basis function approximators for modeling and control,” \textit{in Proceedings of 2009 IEEE International Conference on Control Applications}, St. Petersburg, Russia, Jul. 2009, pp. 1391–1396.

\bibitem{}
L. Fortuna, S. Graziani, A. Rizzo, and M. G. Xibilia, \textit{Soft Sensors for Monitoring and Control of Industrial Processes}, London, U.K.: Springer, 2007


\bibitem{}
Y. Jiang, S. Yin, J. Dong, and O. Kaynak, “A review on soft sensors for monitoring, control, and optimization of industrial processes,” \textit{IEEE Sensors Journal}, vol. 21, no. 11, pp. 12868-12881, Jun. 2021.


\bibitem{}
Z. Chen, Z. Song, and Z. Ge, “Variational inference over graph: Knowledge representation for deep process data analytics,”\textit{ IEEE Transactions on Knowledge and Data Engineering}, vol. 36, no. 6, pp. 2730–2744, Jun. 2024.


\bibitem{}
J. Xie, B. Huang, and S. Dubljevic, “Transfer learning for dynamic feature extraction using variational bayesian inference,”\textit{ IEEE Transactions on Knowledge and Data Engineering}, vol. 34, no. 11, pp. 5524–5535, Nov. 2022.


\bibitem{}
V. Chauhan, A. Tiwari, and S. Arya, “Multi-label classifier based on kernel random vector functional link network,” \textit{in Proceedings of 2020 International Joint Conference on Neural Networks}, Glasgow, United Kingdom: IEEE, Jul. 2020, pp. 1–7.


\bibitem{}
K.-K. Xu, H.-X. Li, and H.-D. Yang, “Kernel-based random vector functional-link network for fast learning of spatiotemporal dynamic processes,”\textit{ IEEE Transactions on Systems Man Cybernetics-Systems}, vol. 49, no. 5, pp. 1016–1026, May 2019.


\bibitem{}
Y. Tian and Y. Zhang, “A comprehensive survey on regularization strategies in machine learning,” \textit{Information Fusion}, vol. 80, pp. 146–166, Apr. 2022.


\bibitem{}
D. Wang and M. J. Felicetti, “Stochastic configuration machines for industrial artificial intelligence,” \textit{arXiv:}2308.13570v6, 2023.


\bibitem{}
M. J. Felicetti and D. Wang, “Stochastic configuration machines: FPGA implementation,” \textit{arXiv:} 2310.19225v1, 2024.



\end{thebibliography}
\end{document}